\documentclass[fullpaper]{nldl}

\renewcommand{\DOI}{12345}

\usepackage[utf8]{inputenc}
\usepackage{graphicx}
\usepackage{authblk}
\usepackage{amsmath}
\usepackage{ctable}
\usepackage{float}
\usepackage{caption}
\usepackage[square, numbers]{natbib}
\usepackage[hyphens]{url}

\newcommand{\tensor}{\mathcal}
\newcommand{\bs}{\boldsymbol}
\input{underbraces}

\usepackage[linesnumbered,ruled,vlined]{algorithm2e}

\usepackage{hyperref}

\title{Compressing CNN Kernels for Videos Using Tucker Decompositions: Towards Lightweight CNN Applications}

\author[1]{Tobias Engelhardt Rasmussen (tenra@dtu.dk)\thanks{Corresponding Author}}
\author[1]{Line H Clemmensen (lkhc@dtu.dk)}
\author[1]{Andreas Baum (andba@dtu.dk)}
\affil[1]{Department of Applied Mathematics and Computer Science, Technical University of Denmark, 2800 Kgs. Lyngby, Denmark}

\date{\vspace{-5ex}}

\begin{document}
\nldlmaketitle

\begin{abstract}  
Convolutional Neural Networks (CNN) are the state-of-the-art in the field of visual computing. However, a major problem with CNNs is the large number of floating point operations (FLOPs) required to perform convolutions for large inputs. When considering the application of CNNs to video data, convolutional filters become even more complex due to the extra temporal dimension. This leads to problems when respective applications are to be deployed on mobile devices, such as smart phones, tablets, micro-controllers or similar, indicating less computational power. 
\newline \indent Kim et al. proposed using a Tucker-decomposition to compress the convolutional kernel of a pre-trained network for images in order to reduce the complexity of the network, i.e. the number of FLOPs \cite{Kim2016}. In this paper, we generalize the aforementioned method for application to videos (and other 3D signals) and evaluate the proposed method on a modified version of the THETIS data set, which contains videos of individuals performing tennis shots. We show that the compressed network reaches comparable accuracy, while indicating a memory compression by a factor of 51. However, the actual computational speed-up (factor 1.4) does not meet our theoretically derived expectation (factor 6). 

\end{abstract}

\section{Introduction}
Neural networks (NNs) are powerful machine learning tools and their use on mobile phones and devices with limited computational power is increasing, thus a general method for making the NNs more efficient is needed. In visual computing the convolutional neural networks (CNNs) have for many years been the state of the art due to their intuitive nature and their ability to learn features in images or videos using filters. A problem with CNNs is that they are computationally heavy. In 2013, Denil et al. concluded that NNs are often over-parametrized and that many of the weights are redundant \cite{Denil2013}. In the same year, Sironi et al. found that a number of convolutional filters can be computed as a linear combination of a smaller number of separable filters thus exploiting the between-filter redundancy \cite{Rigamonti2013}. 

There have been numerous attempts to take advantage of this using different tensor decomposition methods. The usual approach is to compress the weights of a pre-trained network, change the network architecture to fit the subspace projections of the compressed kernel, and fine-tune the new network using the new weights. Attempts to speed up the convolutional operation include Lebedev et al. (2015)\cite{Lebedev2015} using Canonical decomposition (CP), Wang et al. (2016)\cite{Wang2016} using block-term decomposition, Kim et al. (2016)\cite{Kim2016} using Tucker-decomposition, and Jaderberg et al. (2014)\cite{Jaderberg2014} using two original low-rank expansion schemes.
\begin{figure*}
    \centering
    \includegraphics[width=\linewidth]{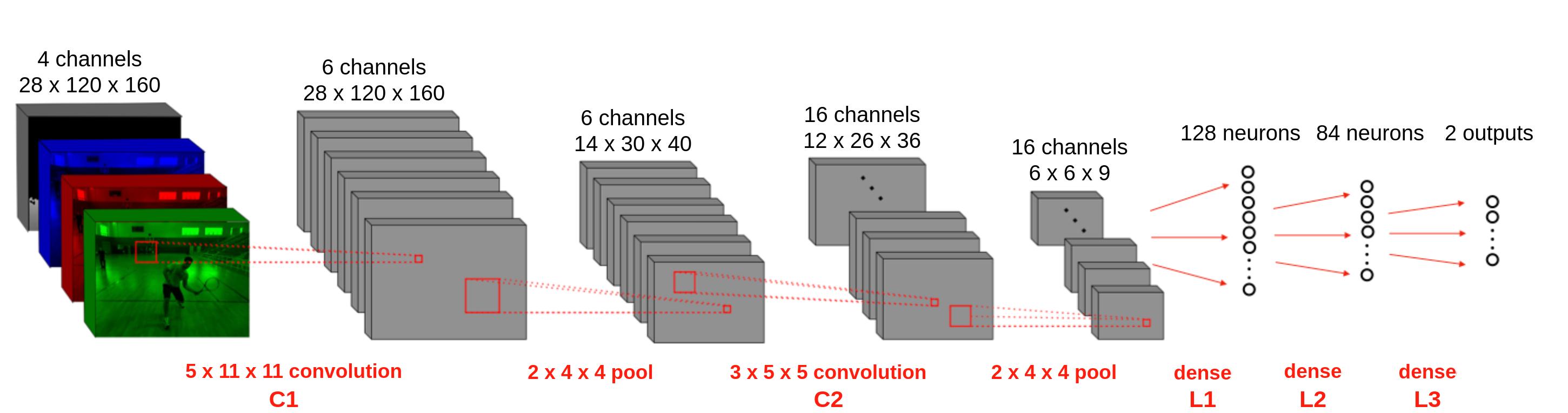}
    \caption{The original architecture used to learn the THETIS data set including layer names}
    \label{fig:architecture}
\end{figure*}

The convolutional operation itself can be understood as the sums of the element-wise multiplications of the filter with the image at a given location. the output of the convolution operation is generated by moving the filter iteratively across the entire image. This makes each parameter responsible for numerous floating point operations (FLOPs), thus making the operation complex and slow. Especially for videos, this becomes an increasing problem due to the additional, temporal dimension.

When classifying videos, rather than images, there are multiple ways of dealing with the added temporal information. The \textit{early fusion} approach can be considered a 3-dimensional convolution because it also includes the temporal dimension of the video. This allows for detection of precise movement and speed because it is calculated \textit{early} at the pixel-level \cite{Karpathy2014}. Early fusion works well for classification of specific actions instead of general activities. The 3D convolutional approach intice us to extend the compression work by Kim et al. 

Kim et al. proposed using a Tucker decomposition to compress both the convolutional kernel and the linear layer in a 2D CNN, thus developing a method able to compress an entire network using the same method, namely the \textit{one-shot whole network compression} algorithm \cite{Kim2016}. This approach yielded promising results, however is not directly applicable for video (3D) convolutions. We propose to generalize the method developed by Kim et al. for images, to also cover video convolutions. We evaluate the method on a modified version of the THETIS data set, \cite{Gourgari2013}, which is described in detail below, and using the architecture given in \autoref{fig:architecture}.

After a short introduction to the data modifications in \autoref{tex:data} the proposed method will be presented in \autoref{tex:method}. The experiments and results will be presented in \autoref{tex:results} and discussed in \autoref{tex:discussion}.

\section{The THETIS Data Set} \label{tex:data}


The THETIS (THree dimEnsional TennIs Shots) action data set
consists of 1980 videos of individuals performing tennis shots. Each of the 12 shot types\footnote{The types include: forehand, backhand, smash, service, volley, etc.\cite{Gourgari2013}} has been performed multiple times by 55 individuals (31 beginners and 24 experts). Each observation consists of an RGB video of a single shot, a depth video (gray-scale), a silhouette video (binary), and both 2D and 3D skeleton videos. The videos have a resolution of $480\times 640$, while they vary in length (approximately 3-7 s). Although the videos are fairly standardized, they still contain a significant amount of variability due to different locations (arena with background noise/changing room), gender (male/female), age (kid-adult), handedness (right/left), and skill set (amateurs/professionals).

\subsection{Modified THETIS data set}
In order to ease the training of the model, the data has been modified as described in the following. First, only two types of shots (forehand and backhand) were selected, making it a binary classification problem. All videos were scaled down by a factor of four in each dimension. The lengths of the videos have been standardized by extracting the same number of frames (14) on either side of a manually assigned time-point reflecting the perceived midway through the shot. Lastly, the RGB video and the black and white depth video have been concatenated, resulting in four input channels for each observation (see \autoref{fig:architecture}). With these modifications the resulting data set contains 327 observations that are all 4-way tensors of size: 4 channels $\times$ 28 frames $\times$ 120 vertical $\times$ 160 horizontal pixels.

\section{Proposed Method} \label{tex:method}
\begin{figure*}
    \centering
    \includegraphics[width=\linewidth]{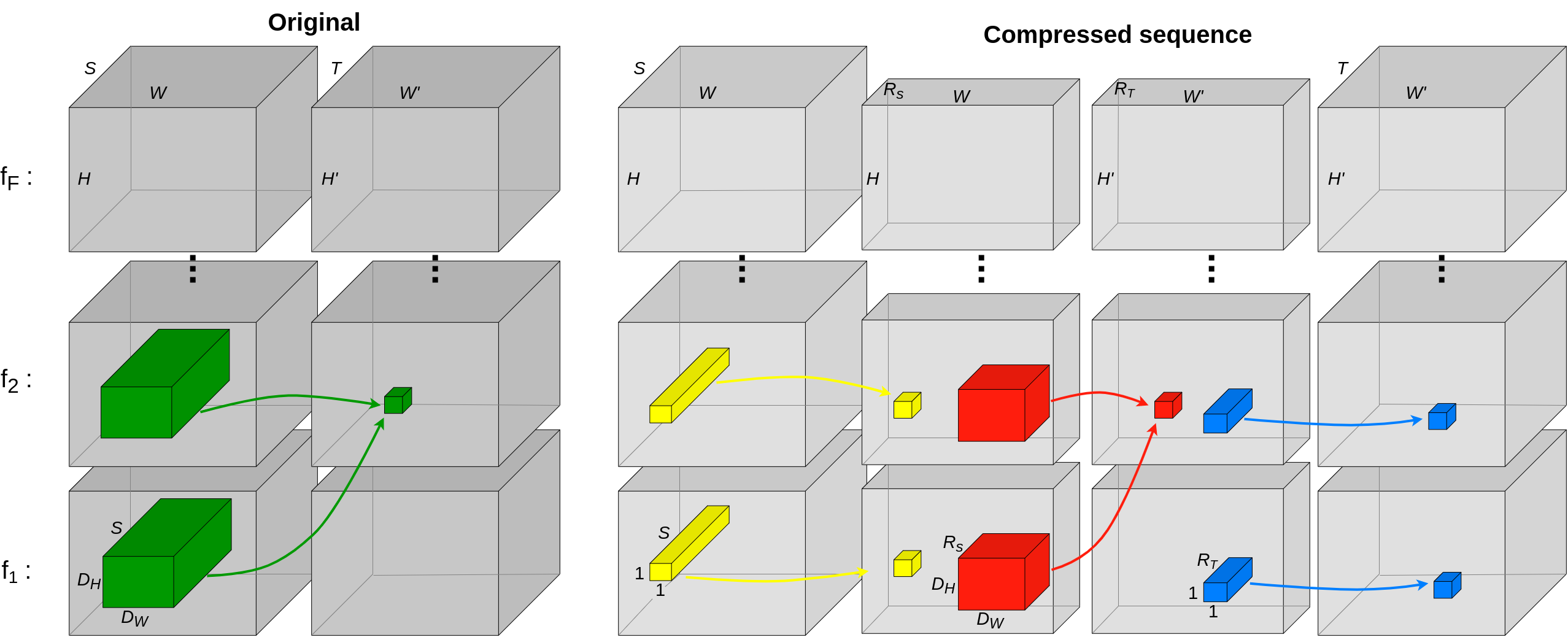}
    \caption{Visualization of the compression of a convolution on a video into a sequence of smaller convolutions based on \cite{Kim2016}. For this visualisation the frame dimension is assumed to have adequate padding such that it does not change. Noticeably, $D_F$ is the time dimension of the kernel $\tensor{K}$, i.e. how many frames it stretches over (two in this illustration).}
    \label{fig:comp_vis}
\end{figure*}
\begin{table*}[h]
\centering
\caption{The number of multiplications and number of parameters needed for the original convolutional operation and the convolutional operation using the compressed kernel, respectively.}
\label{tab:complexity}
\begin{tabular}{l|c|c}
                       & \textbf{Original} & \textbf{Compressed} \\ \hline
\textbf{\# multiplications}         &         $S \cdot T \cdot \Gamma \cdot \Lambda$          &          $S \cdot R_s \cdot \Gamma + R_s \cdot R_t \cdot \Lambda \cdot \Gamma' + R_t \cdot T \cdot \Gamma'$                       \\
 \textbf{\# parameters} &       $S\cdot T \cdot \Gamma $            &        $S\cdot R_s + R_s \cdot R_t \cdot \Gamma + R_t \cdot T$      \end{tabular}
\end{table*}
The proposed method follows the same course of action as in the work done by Kim et al.\cite{Kim2016}, however, the derivation of the compression of the convolutional kernel was described for image input data and will, therefore, be expanded in order to generalize to input data containing 4 dimensions, i.e. an extra video dimension.

\subsection{Tucker-2 Decomposition of the 4D Convolutional Kernel}
We will denote matrices using bold capitals $\bs{X}$ with elements $\bs{X}(i, j) = x_{ij}$ and tensors with $N$ modes using calligraphic letters $\tensor{X}^{d_1\times d_2\times \dots d_N}$ with elements $\tensor{X}(i_1, i_2, \dots, i_N) = x_{i_1i_2\dots i_N}$. 

As described by Kim et al. we will use the Tucker-2 decomposition to compress the input and output channels ($S$ and $T$) in order to exploit the cross-filter redundancy. The spatial and temporal dimensions of the kernel typically do not require compression due to their small size. The convolution of an input tensor $\tensor{X}$ of size  $F\times H \times W \times S$ into an output tensor $\tensor{Y}$ of size $F'\times H' \times W' \times T$ using a video kernel is given by the linear mapping:
\begin{multline}
     \tensor{Y}(f', h', w', t) = \\ \sum_{i=1}^{D_F} \sum_{j=1}^{D_H} \sum_{l=1}^{D_W} \sum_{s=1}^{S} \ \tensor{K}(i, j, l, s, t) \tensor{X}(f_i, h_j, w_l, s)
     \label{eq:3d_convolution}
\end{multline}
where:
\begin{equation}
    f_i = \left(f' - 1\right) \Delta_F + i - P_F    
\end{equation}
\begin{equation}
   h_j =  \left(h' - 1\right) \Delta_H + j - P_H 
\end{equation}
\begin{equation}
   w_l =  \left(w' - 1\right) \Delta_W + l - P_W 
\end{equation}
Here $\Delta_\bullet$ is the stride, $P_\bullet$ is the padding, and $D_\bullet$ is the filter width for the given dimension $F$, $H$, $W$ or $S$. $(f', h', w', t)$ is the position in the output tensor $\tensor{Y}$ and $\tensor{K}$ is the 5-dimensional convolutional kernel, i.e. the stack of $T$ 4-dimensional filters. The tucker-2 decomposition \cite{Moerup2011} of $\tensor{K}$ with respect to the input and output channel dimensions is given by:
\begin{multline}
\tensor{K}(i, j, l, s, t)= \\ \sum_{r_{s}=1}^{R_{s}} \sum_{r_{t}=1}^{R_{t}} \tensor{C}(i, j, l, r_{s}, r_{t}) \ \boldsymbol{U}^{(s)}(s, r_{s}) \  \boldsymbol{U}^{(t)}(t, r_{t})
\end{multline}
Where $\tensor{C}$ is the core of the decomposition of size $D_F\times D_H \times D_W \times R_s \times R_t$ and $\bs{U}^{\bullet}$ is the loading matrix along the input $(s)$ or output $(t)$ dimension, respectively, using rank $R_s$ or $R_t$. Substituting this expression into \eqref{eq:3d_convolution} gives:
\begin{multline}
    \tensor{Y}(f', h', w', t) = \sum_{i=1}^{D_F} \sum_{j=1}^{D_H} \sum_{l=1}^{D_W} \sum_{s=1}^{S} \sum_{r_{s}=1}^{R_{s}} \sum_{r_{t}=1}^{R_{t}} \\ \tensor{C}(i, j, l, r_{s}, r_{t}) \ \boldsymbol{U}^{(s)}(s, r_{s}) \  \boldsymbol{U}^{(t)}(t, r_{t}) \ \tensor{X}(f_i, h_j, w_l, s)
\end{multline}
By rearranging the sums we obtain:
\begin{multline}
    \tensor{Y}(f', h', w', t) = \sum_{r_{t}=1}^{R_{t}} \ \boldsymbol{U}^{(t)}(t, r_{t}) \color{black}\underbracea{\color{white} \underbraceb{ \color{black}\sum_{i=1}^{D_F} \sum_{j=1}^{D_H} \sum_{l=1}^{D_W} \sum_{r_{s}=1}^{R_{s}}}_{\ }} \\ 
    \underbraceb{\tensor{C}(i, j, l, r_{s}, r_{t})  \ \underbrace{\sum_{s=1}^{S} \boldsymbol{U}^{(s)}(s, r_{s}) \ \tensor{X}(f_i, h_j, w_l, s)}_{\large \tensor{Q}(f_i, h_j, w_l, r_s)}}_{\tensor{Q'}(f', h', w', r_t)}
\end{multline}
Initially summing out $s$ yields an intermediate tensor $\tensor{Q}$ of size $F\times H \times W \times R_s$. Subsequently another intermediate tensor $\tensor{Q'}$ of size $F'\times H' \times W' \times R_t$ is obtained by summing out everything except $t$. Using the intermediate tensors, the three linear mappings can be described as in equations \ref{eq:Q}, \ref{eq:Q'} and \ref{eq:Y}.
\begin{multline}
    \tensor{Q}(f, h, w, r_s) = \sum_{s=1}^{S} \boldsymbol{U}^{(s)}(s, r_{s}) \ \tensor{X}(f, h, w, s) \label{eq:Q}
\end{multline}
\begin{multline}
    \tensor{Q'}(f', h', w', r_t) = \sum_{i=1}^{D_F} \sum_{j=1}^{D_H} \sum_{l=1}^{D_W} \sum_{r_{s}=1}^{R_{s}} 
    \\ \tensor{C}(i, j, l, r_{s}, r_{t})   \tensor{Q} (f_i, h_j, w_l, r_s) \label{eq:Q'}
\end{multline}
\begin{multline}
    \tensor{Y}(f', h', w', t) =  \sum_{r_{t}=1}^{R_{t}} \ \boldsymbol{U}^{(t)}(t, r_{t}) \ \tensor{Q}'(f', h', w', r_t) \label{eq:Y}
\end{multline}
These equations correspond to the sequence of convolutions visualized in \autoref{fig:comp_vis}. In detail we have:
\begin{itemize}
    \item Yellow (eq. \ref{eq:Q}) - $1\times 1\times 1$ convolution with $S$ input channels and $R_s$ output channels (the rank of the decomposition).
    \item Red (eq. \ref{eq:Q'}) - $D_F \times D_H \times D_W$ convolution like the original (green), but with $R_s$ input channels and $R_t$ output channels
    \item Blue (eq. \ref{eq:Y}) - $1\times 1\times 1$ convolution with $R_t$ input channels and $T$ output channels
\end{itemize}
The key to the theoretical speed-up is to evaluate the convolution on a smaller tensor using less filters. The first $1\times 1 \times 1$ convolution reduces the number of input channels to $R_s$, hence, the convolution is carried out on $R_s$ input channels and using only $R_t$ filters, and in the end another $1\times 1\times 1$ convolution is bringing the number of channels from $R_t$ to $T$.

The ranks $R_t$ and $R_s$ are estimated using Variational Bayesian Matrix Factorization (VBMF). They are assumed to be small under the assumption that filter parameter redundancy exists. Low ranks are desirable in order to decrease the number of parameters and corresponding number of FLOPs. A complexity analysis is presented below illustrating how ranks can be chosen appropriately.
\begin{table*}[h]
\footnotesize
\centering
\caption{Overview of the layers of the decomposed network resulting after applying the \textit{one-shot whole network compression} scheme to the architecture given in \autoref{fig:architecture} including number of weights and FLOPs for each layer. The number of FLOPs have been used to calculate the theoretical speed up, and the weights have been used to calculate the storage improvement, where $\times x$ means an improvement of $x$ times. The observed time is given as the mean and standard deviation of 500 evaluations. In the bracket, the time for each sub-layer is given.}
\label{tab:res_THETIS_FLOPs}
\begin{tabular}{c|cccccc}
\textbf{Layer}         & Comp. & $ S / R_{in}$ & $ T /R_{out}$ & Weights         & FLOPs                                                                   & CPU time (ms)                                                           \\ \specialrule{0.1em}{.05em}{.05em}
C1     &    & 4             & 6             & 14.5K           & 15609.2M                                                                & $1231.60    \pm 67.06$                                                       \\
Comp     & Tucker2     & 2             & 2             & 2.5K            & \begin{tabular}[c]{@{}c@{}}2618.6M \\ $(=7.5+2600.9+10.2)$\end{tabular} & \begin{tabular}[c]{@{}c@{}}$877.46 \pm 29.27$\\ $(=1.43+872.63+3.40)$\end{tabular} \\
\textit{Impr} &  &             &               & $\times 5.94 $  & $\times 5.96 $                                                          & $\times 1.40 $                                                          \\ \hline
C2    &    & 6             & 16            & 7.2K            & 161.6M                                                                  & $0.82  \pm 0.04      $                                                            \\
Comp     & Tucker2     & 2             & 3             & 526             & \begin{tabular}[c]{@{}c@{}}11.4M \\ $(=0.4+10.8+0.9)$\end{tabular}      & \begin{tabular}[c]{@{}c@{}}$0.81\pm 2.09$ \\ $(=0.17+0.39+0.25)$\end{tabular}      \\
\textit{Impr} &     &           &               & $\times 13.72 $ & $\times 14.2$                                                           & $\times 1.02 $                                                          \\ \hline
L1   &   & 16            & 128           & 663.7K          & 1.3M                                                                    & $0.70  \pm 0.05    $                                                              \\
Comp     & Tucker2     & 4             & 7             & 10.1K           & \begin{tabular}[c]{@{}c@{}}60.0K \\ $(=40.2+18.1+1.7)$\end{tabular}     & \begin{tabular}[c]{@{}c@{}}$0.36 \pm 0.55$\\ $(=0.11+0.17+0.08)$\end{tabular}      \\
\textit{Impr} &       &         &               & $\times 65.32 $ & $\times 22.12 $                                                         & $\times 1.97 $                                                          \\ \hline
L2    &     & 128           & 84            & 10.8K           & 21.5K                                                                   & $0.08    \pm 0.51                $                                                \\
Comp    & Tucker1      & -             & 1             & 296             & \begin{tabular}[c]{@{}c@{}}423 \\ $(=255+168)$\end{tabular}             & \begin{tabular}[c]{@{}c@{}}$0.07 \pm 0.04$\\ $(=0.03+0.03)$\end{tabular}           \\
\textit{Impr} &   &             &               & $\times 36.61 $ & $\times 50.84 $                                                         & $\times 1.15 $                                                          \\ \hline
L3      &   & 84            & 2             & 170             & 336                                                                     & $0.03 \pm 0.002$                                                                   \\
Comp     & -     & -             & -             & 170             & 336                                                                     & $0.03 \pm 0.002$                                                                    \\
\textit{Impr} & & \textit{}     & \textit{}     & $\times 1.00$   & $ \times 1.00 $                                                         & $\times1.03$                                                           \\ \specialrule{0.1em}{.05em}{.05em}
Total      &    &               &               & 696.4K          & 15772.1M                                                                & $1238.56    \pm 67.33$                                                                    \\
Comp      &    &               &               & 13.6K           & 2630.1M                                                                 & $884.02 \pm 29.91$                                                                       \\
\textit{Impr} &  &&               & $ \times 51.22$ & $\times 6.0 $                                                           & $\times 1.40$                                                                        \\ \hline
\end{tabular}
\end{table*}

\subsubsection{Complexity Analysis} \label{tex:complexity}
In this section we discuss the number of parameters and multiplications (proportional to FLOPs) required to perform a convolution operation on an input video of size $F\times H\times W$ into an output video of size $F'\times H'\times W'$ using a kernel of size $D_F\times D_H\times D_W\times S\times T$. In the following we define $\Gamma = F \cdot H \cdot W$, \ \ $\Gamma' = F' \cdot H' \cdot W'$ and $\Lambda = D_F \cdot D_H \cdot D_W$ (number of pixels per channel in the input image, output image, and kernel respectively).
The number of multiplications and parameters needed to perform the original vs. compressed mapping are given in \autoref{tab:complexity}. The ratios between the original and the compressed version for, both, the number of multiplications and the number of parameters are bound by the ratio between the product of the original in- and output dimensions; and the product of the ranks of the compressed kernel, i.e. $\frac{S\cdot T}{R_s\cdot R_t}$.


\subsection{One-Shot Network Compression}
The \textit{one-shot whole network compression} algorithm reported by Kim et al. \cite{Kim2016} is used with minor modifications. VBMF is applied for rank selection in a given mode. Prior to this, the kernel is matricized accordingly \cite{Moerup2011}. 
For the compression part it is noteworthy to mention that the Tucker-2 decomposition is used on all convolutional layers but the first as well as on the first linear layer (i.e. C2$\dots$ C$N_C$ and L1)\footnote{C$\bullet$ and L$\bullet$ are convolutional respectively linear layers in the architecture with $N_C$ convolutional layers followed by $N_L$ linear layers}. Tucker-1 is used on the remaining layers (i.e. C1 and L2$\dots$ L$N_L$). To allow for Tucker-2 decomposition of the first linear layer, it is compressed treating it as a convolutional layer. The full algorithm is provided in \autoref{alg:one_shot_compression}.

\begin{algorithm*}[ht] \caption{One-Shot Tucker Compression of a Video CNN} \label{alg:one_shot_compression}
\small
    $\texttt{net} \gets$ Define appropriate CNN with convolutional layers C1, C2, \dots, CN$_C$ and linear layers L1, L2, \dots,  LN$_L$\;
    \For{\texttt{layer} in \{C2,\dots CN$_C$, L1\}}{
        $\tensor{K}_{\texttt{layer}} \leftarrow$ Take out weight kernel\;
        $R_s, R_t \leftarrow$ Choose ranks using VBMF\;
        $\tensor{G}, \ \ \bs{U}^{(s)}, \bs{U}^{(t)} \leftarrow$ Compress $K_{\texttt{layer}}$ using Tucker-2 on in/out dimensions\;
        $\texttt{comp1}, \texttt{comp2}, \texttt{comp3} \leftarrow$ Define new layers\;
        $\tensor{K}_{\texttt{comp1}} \leftarrow \bs{U}^{(s)}$ \tcp*{$1\times 1\times 1$ convolution}
        $\tensor{K}_{\texttt{comp2}} \leftarrow \tensor{G}$ \tcp*{$D_F\times D_H\times D_W$ convolution}
        $\tensor{K}_{\texttt{comp3}} \leftarrow \bs{U}^{(t)}$ \tcp*{$1\times 1\times 1$ convolution}
        $\bs{b}_{\texttt{comp3}} \leftarrow \bs{b}_{\texttt{layer}}$ \tcp*{Add bias to last layer}
        $\texttt{layer} \leftarrow \texttt{Sequential(comp1, comp2, comp3)}$
    }
    \For{\texttt{layer} in \{C1, L2,\dots, LN$_L$\}}{
        $\tensor{K}_{\texttt{layer}} \leftarrow$ Take out weight kernel\;
        $R \leftarrow$ Choose rank using global analytical VBMF\;
        $\tensor{G}, \ \ \bs{U} \leftarrow$ Compress $K_{\texttt{layer}}$ using Tucker-1 on out dimensions\;
        $\texttt{comp1}, \texttt{comp2} \leftarrow$ Define new layers\;
        $\tensor{K}_{\texttt{comp1}} \leftarrow \tensor{G}$ \tcp*{$D_F\times D_H\times D_W$ convolution}
        $\tensor{K}_{\texttt{comp2}} \leftarrow \bs{U}$ \tcp*{$1\times 1\times 1$ convolution}
        $\bs{b}_{\texttt{comp2}} \leftarrow \bs{b}_{\texttt{layer}}$ \tcp*{Add bias to last layer}
        $\texttt{layer} \leftarrow \texttt{Sequential(comp1, comp2)}$
        }
    train \texttt{layer} on training data \tcp*{fine-tune the compressed network}
\end{algorithm*}
\section{Experiments and Results} \label{tex:results}
The network architecture used to train the video data is illustrated in \autoref{fig:architecture}. A hyper-parameter search was conducted using 5-fold cross validation due to the limited number of observations. After applying the \textit{one-shot whole network compression} scheme, the compressed network was fine-tuned reaching the same accuracy (90.9\%) as the original network; in very few epochs. The resulting network is shown in \autoref{tab:res_THETIS_FLOPs} along with the theoretical speed-up and storage improvements calculated from the number of FLOPs and parameters, respectively. The observed speed-up computed using the \texttt{profiler}-module in \texttt{torch} \cite{pytorch} is also given.

The total elapsed time also includes delays caused by non-linearities, transformations, and poolings, hence is not a direct sum of the individual layers. These extra computations are the same for both models, and therefore give a more truthful value for the speed-up of the entire model. 

From \autoref{tab:res_THETIS_FLOPs} it is clear that the improvements in terms of weights and FLOPs are substantial with an overall storage improvement of more than 51 times, and a theoretical speed-up of 6 times. This is not the case for the actual computation time that, even though it has decreased by a factor 1.4, does not meet that of the expected in any case.

\section{Discussion} \label{tex:discussion}
From the results, it is clear that the observed speed-up does not meet the expected. The key operation in the given scheme is the $1\times 1\times 1$ convolution that is used to bring down the dimensions for the actual convolution allowing it to run much faster. In the work by Kim et al. \cite{Kim2016}, they conclude that the lack of observed speed-up is caused by the cache-inefficiency of the $1\times 1$ convolution which can be confirmed by the results achieved here. For example for the second convolution, the $1\times 1\times 1$ convolutions take up 31\% and 22\% of the time while they only correspond to 4\% and 8\% of the FLOPs, respectively. For the first, bigger convolution the observed speed-up is greater hence it seems that the proposed scheme works better for bigger layers. Kim et al. also report greater observed speed-ups when running on smaller devices with less computational power, which makes it seem that the inefficiency of the $1\times 1\times 1$ convolution is more pronounced with more computational power, hence the proposed scheme is more appropriate for low-power applications. 

\texttt{PyTorch} is well optimised in terms of evaluation due to its ability to make use of all computational power accessible \cite{pytorch}. This means that it is excellent at performing large operations quickly, and will likely be prone to penalties caused by additional yet smaller layers. This fact implies that caution should be taken when choosing optimal ranks as ranks too big will result in too many FLOPs according to section \ref{tex:complexity}, but choosing ranks too small will increasingly result in penalties towards evaluation optimisation performed by \texttt{PyTorch}. 

We believe that the theoretical speed-up could be approached, but that it would require the hardware to have a more linear relation between execution time and the number of operations required. For future work, this could be investigated by use of small devices with limited computational power (that have no optimization of the calculations by e.g. parallelization) or by defining the calculations outside of the PyTorch framework and potentially translating it into C. 

\bibliographystyle{abbrvnat}
\bibliography{references}
\end{document}